\documentclass[runningheads]{llncs}
\usepackage{graphicx}
\usepackage{amsmath,amssymb} 
\usepackage{color}

\usepackage{caption}
\usepackage{subcaption}
\usepackage{bm}
\usepackage{multirow}
\usepackage{hyperref}
\usepackage{stfloats}
\usepackage{verbatim}
\usepackage{makecell}

\newcommand{\matr}[1]{\bm{#1}}
\begin{document}

\newcommand{\point}{
    \raise0.7ex\hbox{.}
    }


\pagestyle{headings}

\mainmatter

\title{Image Patch Matching Using Convolutional Descriptors with Euclidean Distance} 

\titlerunning{Image Patch Matching Using Convolutional Descriptors with Euclidean Distance} 

\authorrunning{Iaroslav Melekhov\inst{1}, Juho Kannala\inst{1}, Esa Rahtu\inst{2}} 

\author{Iaroslav Melekhov\inst{1}, Juho Kannala\inst{1}, Esa Rahtu\inst{2}} 
\institute{Department of Computer Science, Aalto University, Finland \newline
\and Center for Machine Vision Research, University of Oulu, Finland}

\maketitle

\begin{abstract}
In this work we propose a neural network based image descriptor suitable for image patch matching, which is an important task in many computer vision applications. Our approach is influenced by recent success of deep convolutional neural networks (CNNs) in object detection and classification tasks. We develop a model which maps the raw input patch to a low dimensional feature vector so that the distance between representations is small for similar patches and large otherwise. As a distance metric we utilize $L_2$ norm, i.e.~Euclidean distance, which is fast to evaluate and used in most popular hand-crafted descriptors, such as SIFT. According to the results, our approach outperforms state-of-the-art $L_2$-based descriptors and can be considered as a direct replacement of SIFT. In addition, we conducted experiments with batch normalization and histogram equalization as a preprocessing method of the input data. The results confirm that these techniques further improve the performance of the proposed descriptor. Finally, we show promising preliminary results by appending our CNNs with recently proposed \emph{spatial transformer networks} and provide a visualisation and interpretation of their impact.
\end{abstract}

\section{Introduction} \label{sec:intro}

Finding correspondences between image regions (patches) is a key factor in many computer vision applications. For example, structure-from-motion, multi-view reconstruction, image retrieval and object recognition require accurate computation of local image similarity. Due to importance of these problems various descriptors have been proposed for patch matching with the aim of improving accuracy and robustness. Many of the most widely used approaches, like SIFT \cite{SIFT} or DAISY \cite{DAISY} descriptors, are based on hand-crafted features and have limited ability to cope with negative factors (occlusions, variation in viewpoint etc.) making a search of similar patches more difficult. Recently, various methods based on supervised machine learning have been successfully applied for learning patch descriptors \cite{BrownPatches,Trzcinski1,Trzcinski2,SimonyanDescriptor}. These methods significantly outperform hand-crafted approaches and inspire our research.

During recent years, neural networks have achieved great success in object classification \cite{Krizhevsky} and other computer vision problems. Specifically, methods based on Convolutional Neural Network (CNN) have showed significant improvements over previous state-of-the-art recognition and object detection approaches. Influenced by these works, we aim to create a CNN-based  
 discriminative descriptor for patch matching task. In contrast to~\cite{Zagoruyko,Matchnet} where the representations of two patches are compared using a set of fully connected layers, we utilize \textit{Euclidean distance} as a metric of similarity. The same metric is used in one of the most popular and applicable descriptor, SIFT. Therefore, our approach can be considered as a direct alternative to SIFT and similar techniques can be used for fast matching and indexing of descriptors as with SIFT. We utilize labeled patch pairs to learn the descriptor so that Euclidean distance ($L_2$ norm) between patches in the feature space is small for similar patches and large otherwise. This is analogous to face-verification problem where Siamese structure~\cite{lecunSiameseNetwork} has been utilized to predict whether the persons illustrated in an input image pair are the same or not. 

\begin{figure}[t!]
\centering
	\begin{subfigure}{0.49\linewidth}
	\centering
		\includegraphics[scale=0.3]{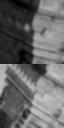}\hspace{0.4em}%
		\vspace{0.1em}
		\includegraphics[scale=0.3]{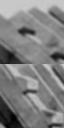}\hspace{0.4em}%
		\vspace{0.1em}
		\includegraphics[scale=0.3]{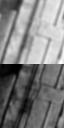}\hspace{0.4em}%
		\vspace{0.1em}
		\includegraphics[scale=0.3]{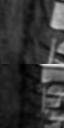}\hspace{0.4em}%
		\vspace{0.1em}
		\includegraphics[scale=0.3]{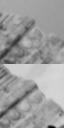}\hspace{0.4em}%
		\vspace{0.1em}
		\includegraphics[scale=0.3]{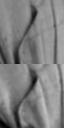}\hspace{0.4em}%
		\vspace{0.1em}
		\includegraphics[scale=0.3]{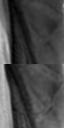}\hspace{0.4em} %
		\vspace{0.1em}
		\includegraphics[scale=0.3]{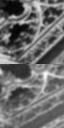}\hspace{0.4em}%
		\vspace{0.1em}
		\includegraphics[scale=0.3]{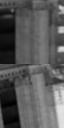}\hspace{0.4em}%
		\vspace{0.1em}
		\includegraphics[scale=0.3]{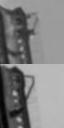}\hspace{0.4em}%
		\vspace{0.1em}
		\includegraphics[scale=0.3]{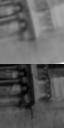}\hspace{0.4em}%
		\vspace{0.1em}
		\includegraphics[scale=0.3]{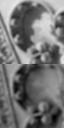}\hspace{0.4em}%
		\vspace{0.1em}
		\includegraphics[scale=0.3]{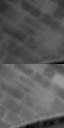}\hspace{0.4em}%
		\vspace{0.1em}
		\includegraphics[scale=0.3]{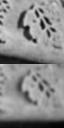}\hspace{0.4em}%
		\vspace{0.1em}
		\includegraphics[scale=0.3]{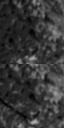}\hspace{0.4em}%
		\includegraphics[scale=0.3]{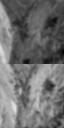}\hspace{0.4em}%
		\includegraphics[scale=0.3]{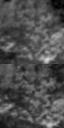}\hspace{0.4em}%
		\includegraphics[scale=0.3]{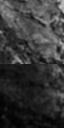}\hspace{0.4em}%
		\includegraphics[scale=0.3]{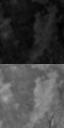}\hspace{0.4em}%
		\includegraphics[scale=0.3]{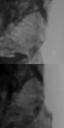}\hspace{0.4em}%
		\includegraphics[scale=0.3]{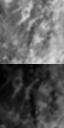}\hspace{0.4em}%
		\caption{positive pairs}
	\end{subfigure}
	\begin{subfigure}{0.49\linewidth}
	\centering
		\includegraphics[scale=0.3]{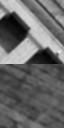}\hspace{0.3em}%
		\vspace{0.1em}
		\includegraphics[scale=0.3]{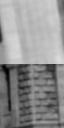}\hspace{0.3em}%
		\vspace{0.1em}
		\includegraphics[scale=0.3]{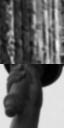}\hspace{0.3em}%
		\vspace{0.1em}
		\includegraphics[scale=0.3]{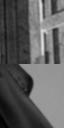}\hspace{0.3em}%
		\vspace{0.1em}
		\includegraphics[scale=0.3]{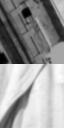}\hspace{0.3em}%
		\vspace{0.1em}
		\includegraphics[scale=0.3]{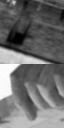}\hspace{0.3em}%
		\vspace{0.1em}
		\includegraphics[scale=0.3]{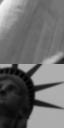}\hspace{0.3em}%
		\vspace{0.1em}
		\includegraphics[scale=0.3]{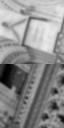}\hspace{0.3em}%
		\vspace{0.1em}
		\includegraphics[scale=0.3]{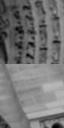}\hspace{0.3em}%
		\vspace{0.1em}
		\includegraphics[scale=0.3]{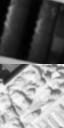}\hspace{0.3em}%
		\vspace{0.1em}
		\includegraphics[scale=0.3]{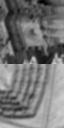}\hspace{0.3em}%
		\vspace{0.1em}
		\includegraphics[scale=0.3]{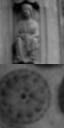}\hspace{0.3em}%
		\vspace{0.1em}
		\includegraphics[scale=0.3]{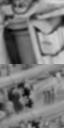}\hspace{0.3em}%
		\vspace{0.1em}
		\includegraphics[scale=0.3]{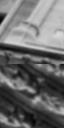}\hspace{0.3em}%
		\vspace{0.1em}
		\includegraphics[scale=0.3]{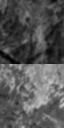}\hspace{0.3em}%
		\includegraphics[scale=0.3]{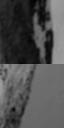}\hspace{0.3em}%
		\includegraphics[scale=0.3]{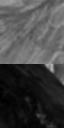}\hspace{0.3em}%
		\includegraphics[scale=0.3]{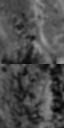}\hspace{0.3em}%
		\includegraphics[scale=0.3]{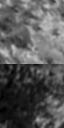}\hspace{0.3em}%
		\includegraphics[scale=0.3]{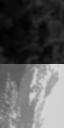}\hspace{0.3em}%
		\includegraphics[scale=0.3]{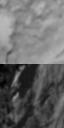}\hspace{0.3em}%
		\caption{negative pairs}
	\end{subfigure}
	\vspace{-3mm}
\caption{Randomly picked matching (i.e.\ positive) and non-matching (i.e.\ negative) patch pairs of Multi-view Stereo Correspondence (MSC) dataset \cite{BrownPatches} which consists of three subsets: Liberty (top row), Notredame (middle row) and Yosemite (bottom row). The matching patches represent the same 3D structure so that their orientation, scale and location are roughly corresponding but there are still significant variations in viewpoint and illumination. The non-matching patches represent different 3D points and therefore they usually have quite different texture and appearance.
} \label{fig:DatasetsExample}
\end{figure}

For training and evaluation of the proposed descriptor we utilize Multi-view Stereo Correspondence (MSC) dataset~\cite{BrownPatches}, which is illustrated in Fig.~\ref{fig:DatasetsExample} and consists of more than 1.5M grayscale patches. The dataset consists of pairs of matching and non-matching patches extracted from images of the Statue of Liberty, Notredame and Half Dome (Yosemite) by using Difference of Gaussian (DoG) interest point detector and matched by utilizing the respective 3D multi-view reconstructions computed from the images \cite{BrownPatches}. In detail, corresponding interest points were found by mapping between images using the dense stereo depth maps computed by the multi-view stereo algorithm of \cite{Goesele2007} based on the initial point cloud reconstructions by \cite{Snavely2008}. Pairs of patches corresponding to the same 3D point are defined to be matching (i.e.~\emph{positive} or \emph{similar} pairs in our terminology) if they also originate from DoG interest points detected with sufficiently similar scale and orientation \cite{BrownPatches}. Pairs of patches sampled from different 3D points are non-matching (i.e.~\emph{negative} or \emph{dissimilar}). In summary, as illustrated in Fig.~\ref{fig:DatasetsExample}, the matching pairs represent the same 3D structure with roughly correct geometric alignment so that their appearances are similar whereas the negative pairs typically have different texture and dissimilar appearance.

In this work, we conduct multiple experiments with preprocessing of raw patches and demonstrate that histogram equalization as well as batch normalization significantly improve the accuracy of the proposed descriptor. 

We also explore different types of descriptor architectures evaluating their performance on MSC dataset. Our experimental evaluation shows that the proposed model outperforms recent state-of-the-art $L_2$-based approaches. In addition, we investigate the use of spatial transformer networks \cite{stn} in the patch matching problem. 

The paper is organized as follows. Section~\ref{sec:relatedWork} presents related work focusing on patch matching problem. Section~\ref{sec:neuralDescriptor} describes the proposed method of finding corresponding patches, discusses an architecture of the descriptor, objective function and details of data preprocessing. Section~\ref{sec:experiments} presents the experimental pipeline and performance on the MSC dataset. In the end of this paper we summarize our results and point some directions of future work.

\section{Related work} \label{sec:relatedWork}

Local image descriptors have been widely used in finding similar and dissimilar regions in images. Nowadays, the trend has changed from hand-crafted and carefully-designed methods (SIFT~\cite{SIFT} or DAISY~\cite{DAISY}) to a new generation of learned descriptors including unsupervised and supervised techniques like boosting~\cite{Trzcinski1}, convex optimization~\cite{SimonyanDescriptor} and Linear Discriminant Analysis (LDA) ~\cite{BrownPatches,LDAHash}.

In our approach, however, we propose a descriptor based on deep convolutional neural networks (CNN) with batch normalization units accelerating learning and convergence. The first papers which utilized CNN based representations for finding matching image patches were \cite{Jahrer2009} and \cite{Osendorfer}. More recently, {\v Z}bontar and LeCun~\cite{ZbontarLeCun} proposed a method for comparing image patches in order to extract stereo depth information. Their method is based on using convolutional networks minimizing a hinge loss function and showed the best performance on KITTI stereo evaluation dataset~\cite{Geiger2013IJRR}. However, as that approach operates on very small patches ($9\times9$ pixels), it restricts the area of applicability. 

In addition, one recent related paper is \cite{cvpr2015geolocalization}, which utilizes Siamese network architecture for the challenging problem of matching street-level and aerial images. In contrast to our work, \cite{cvpr2015geolocalization} concentrates on matching entire images in a specific application, i.e.\ ground-to-aerial geolocalization. Their approach is therefore not directly applicable in tasks where local features are currently used and it does not allow replacing or comparing with SIFT. Moreover, in their work the length of the proposed descriptor is significantly larger ($4,096$) than that of SIFT and our representation ($128$).

Recent approaches \cite{Zagoruyko,Matchnet,simo2015deepdesc} propose CNN descriptors trained with two-branch (Siamese) architecture which significantly exceed the accuracy of hand-crafted descriptors. However, in contrast to SIFT, in \cite{Zagoruyko,Matchnet} the feature representations of input patches are compared by a set of fully connected layers (match network) that learns a complex comparison metric. Nevertheless, Zagoruyko et al.~\cite{Zagoruyko} and Simo-Serra et al.~\cite{simo2015deepdesc} also conducted experiments in which the match network was replaced with Euclidean distance metric between the outputs of two branches and, hence, they are the closest works to ours. The implementation of~\cite{simo2015deepdesc} is not yet publicly available. Thus, in order to compare performance, we reproduced the network architecture of \cite{simo2015deepdesc} and evaluated it using the standard protocol. The results show that our network architecture outperforms those of \cite{Zagoruyko,simo2015deepdesc}. More detailed comparison is presented in Sec.~\ref{ssec:architectureAndLearning}.

\section{Neural Descriptor}\label{sec:neuralDescriptor}

\begin{figure}[t!]
\begin{center}
     \includegraphics[width=.6\textwidth]{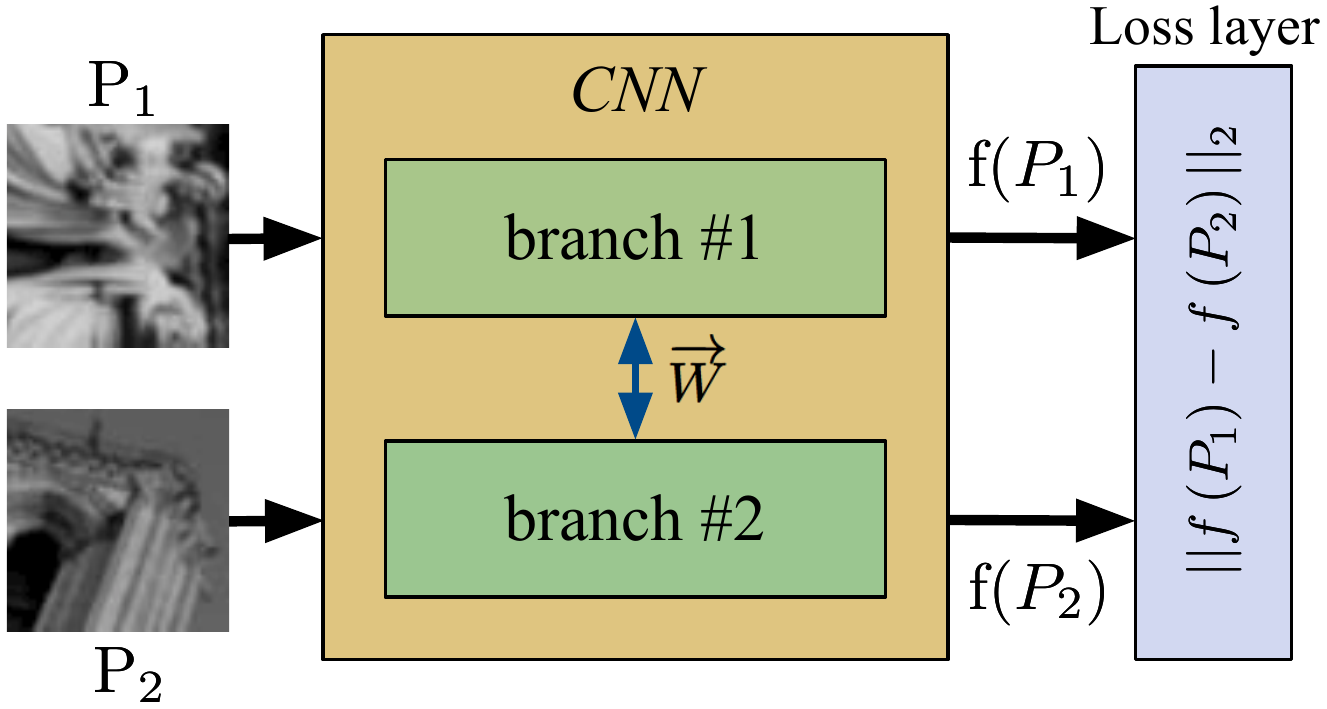}
     \caption{Schematic illustration of the proposed descriptor based on Siamese architecture~\cite{lecunSiameseNetwork}. A pair of patches ($P_1$, $P_2$) is propagated through the network consisting of two identical branches and sharing the same set of weights ($\bm{W}$). Feature representations of patches ($f\left(P_1\right)$, $f\left(P_2\right)$) are extracted from the last layer of each branch separately and Euclidean distance is computed between them. Our objective is to learn a descriptor that minimizes the distance between similar pairs of patches and maximizes it for dissimilar pairs. It is important to note that at test time (i.e.\ after learning) the feature descriptor $f$ can be computed independently for each individual patch since both branches are identical.}
\label{fig:pipelineScheme}
\end{center}
\end{figure}

Our goal is to construct a system that efficiently distinguishes matching (similar) and non-matching (dissimilar) patches. To do this, we propose a method based on a deep convolutional neural network. As shown in Fig.~\ref{fig:pipelineScheme}, the model consists of two identical branches that share the same set of weights and parameters. Patches $P_1$ and $P_2$ are fed into branches and propagated through the model separately. The main objective of a proposed network is to map the raw patches to a low dimensional feature space so that the $L_2$  distance between pairs is small if the patches are similar and large otherwise. The same distance measure ($L_2$ distance) is usually applied also for matching hand-crafted descriptors.

The following section describes the proposed loss function and how it can be used in our approach.

\subsection{Loss Function and Data Preprocessing}\label{ssec:subObjectiveFunction}

To optimize the proposed network, we have to use a loss function which is capable to distinguish similar (positive) and dissimilar (negative) pairs. More precisely, we train the weights of the network by using a loss function which encourages similar examples to be close, and dissimilar ones to have Euclidean distance larger or equal to a margin $m$ from each other. In contrast to \cite{Zagoruyko,simo2015deepdesc}, which utilize hinge embedding loss \cite{hingeLoss}, we use margin-based contrastive loss \cite{lecunContrastiveLoss} defined as follows:

\begin{equation}\label{eq:lossFunction}
  \mathcal{L}\left(P_1, P_2, l \right) = \frac{1}{2}lD^2+ \frac{1}{2}\left(1-l \right)\left\{\max\left(0, m-D\right)\right\}^2
\end{equation}
where $\mathit{l}$ is a binary label which selects whether the input pair consisting of patch $P_1$ and $P_2$ is a positive ($l = 1$) or negative ($l = 0$), $\mathit{m >}$ 0 is the margin for negative pairs and
$D = \left\|f(P_1) - f(P_2)\right\|_2$ is the Euclidean Distance between feature vectors $f(P_1)$ and $f(P_2)$ of input images $P_1$ and $P_2$.

\begin{figure}[t!]
\centering
		\centering
		\includegraphics[scale=0.5]{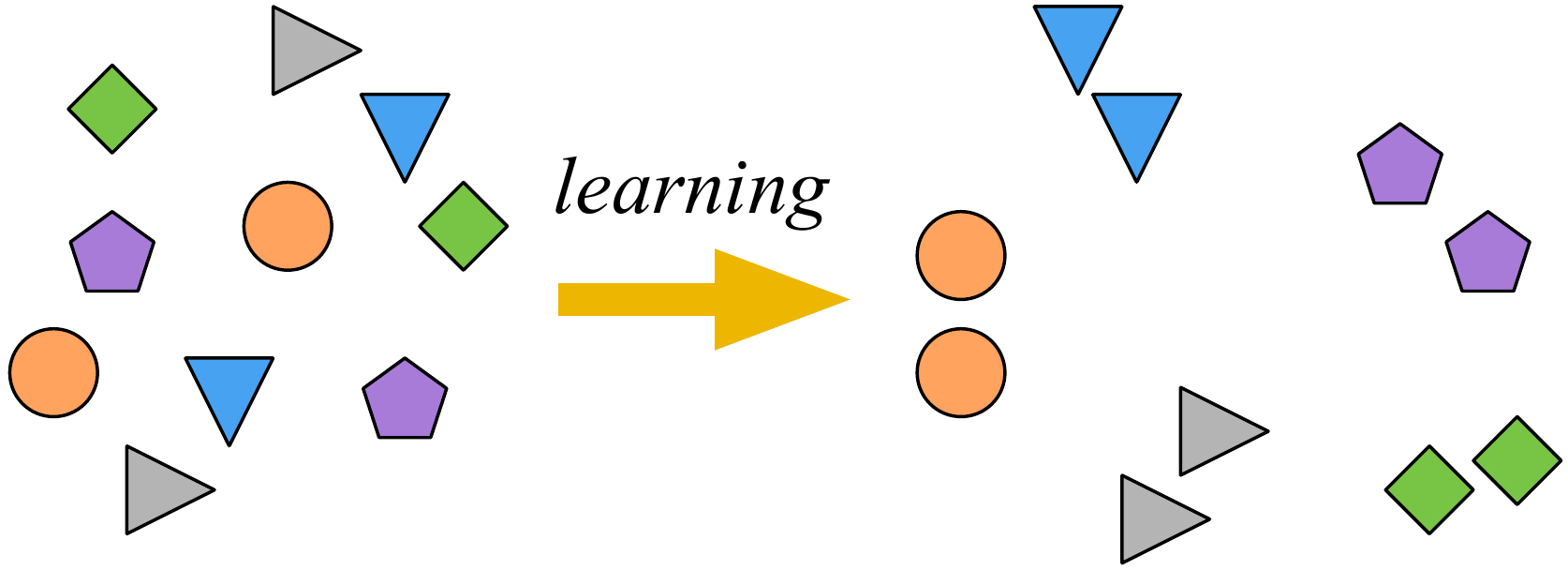}
		\caption{Contrastive loss minimizes the distance between positive patch pairs (elements with the same color and shape) and maximizes otherwise.}\label{fig:learningGoal}
\end{figure}

Dissimilar pairs contribute to the loss function only if their distance is smaller than the margin $m$. The idea of learning is schematically illustrated in Fig.~\ref{fig:learningGoal}. The loss function encourages matching patches (elements with the same color and shape) to be close in feature space while pushing non-matching pairs apart. Obviously, negative pairs with a distance larger than margin would not contribute to the loss (second part of~(\ref{eq:lossFunction})). Thus, setting margin to too small value would lead to optimizing the objective function only over the set of positive pairs and, as a result, would hamper learning.

\begin{figure}[t!]
	\begin{subfigure}{0.49\linewidth}
		\centering
		\includegraphics[scale=0.5]{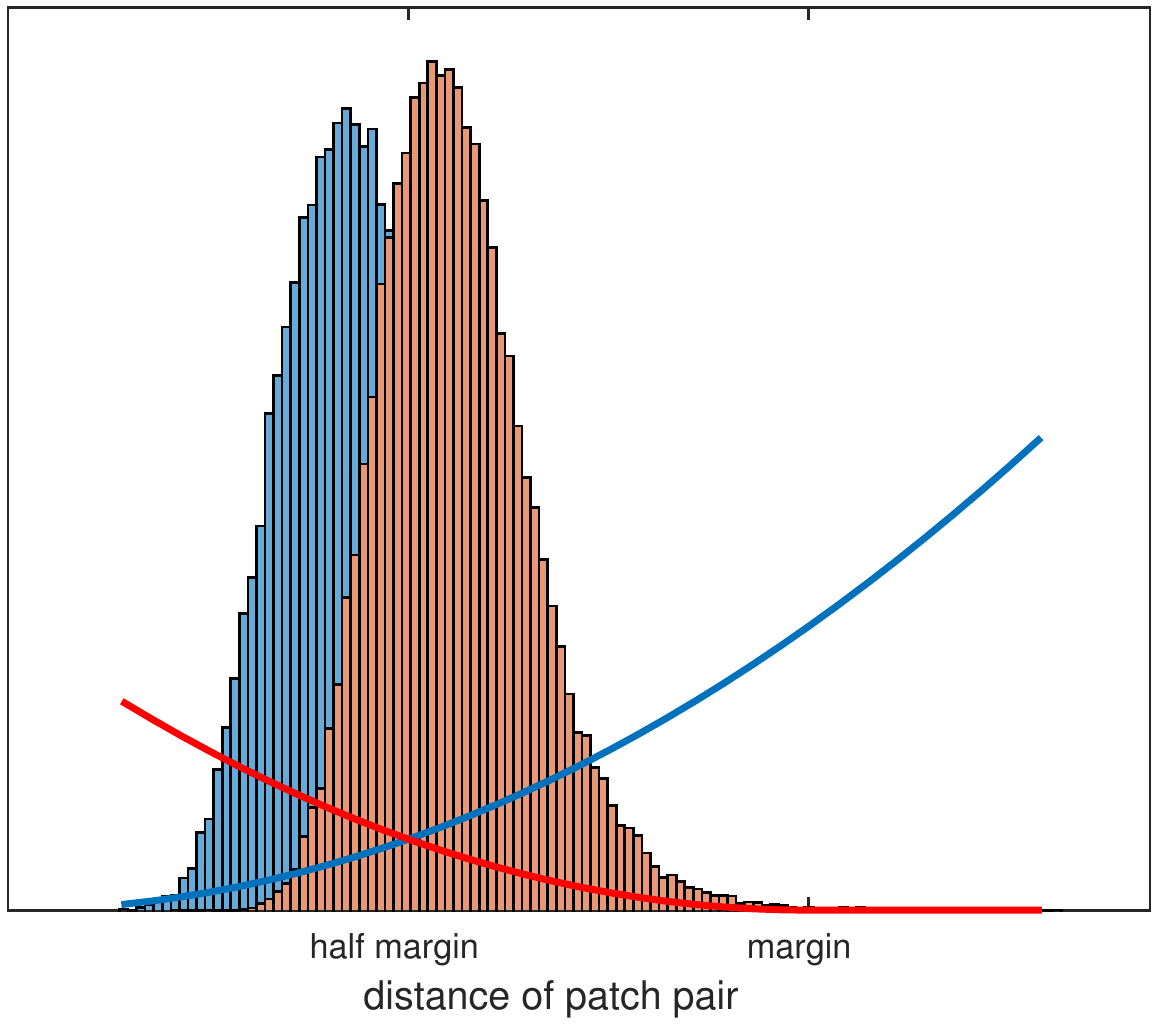}
		\caption{Before learning}\label{sub:distributionsBeforeLearning}
	\end{subfigure}\hspace{0.3em}%
	\begin{subfigure}{0.49\linewidth}
		\centering
		\includegraphics[scale=0.5]{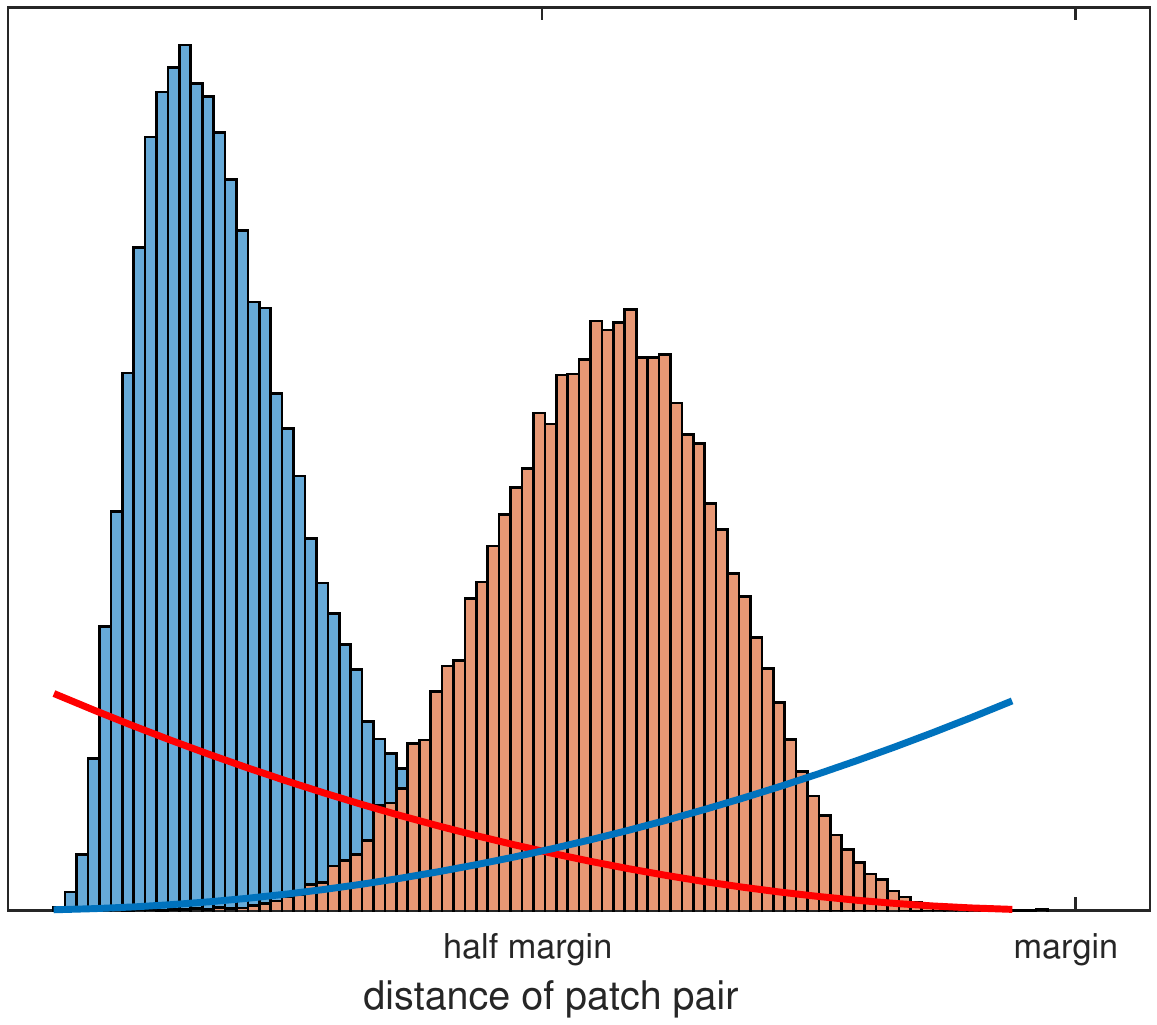}
		\caption{After learning}\label{sub:distributionsAfterLearning}
	\end{subfigure}
\caption{The distributions of feature distances $D$ for positive (blue) and negative (brown) patch pairs of Notredame test dataset before (left) and after (right) training on Liberty patches of MSC dataset. Learning decreases distances of positive pairs and increases distances of negative pairs. The blue curve is the loss for positive pairs, i.e.\ $D^2/2$ in \eqref{eq:lossFunction}, and the red curve is the loss for negative pairs, i.e.\ $(\max(0,m-D))^2/2$. The curves intersect at $m/2$. } \label{fig:learningResults}
\end{figure}

To demonstrate what has been learned by our proposed descriptor, we illustrate the histogram of pairwise Euclidean distances of patch pairs of test set both before and after training in Fig.~\ref{fig:learningResults}. The blue and brown bars represent pairwise distances of positive and negative pairs, respectively. It can clearly be seen that the training process of the descriptor on patch pairs effectively pushes non-matching pairs away and pulls matching pairs together. In the very beginning, the distributions of positive and negative pairs are grouped at the intersection of the blue (penalty for similar pairs~(\ref{eq:lossFunction})) and the red (penalty for dissimilar pairs) curves in Fig.~\ref{sub:distributionsBeforeLearning}. We experimentally verified that for efficient training the margin value should be set to twice the average Euclidean distance between features of training patch pairs before learning.

\paragraph{Data Preprocessing and Augmentation.} Data preprocessing plays an important role in machine learning algorithms. However, in practice it is hard to say in advance which preprocessing technique is helpful for achieving best performance. Here we calculate mean and standard deviation of pixel's intensities over the whole training dataset and use them to normalize intensity value of every pixel in the input grayscale patch. In addition, analysing raw patches in MSC dataset, we noticed that there are a lot of pairs where patches have significantly different contrast. To adjust patch intensities we apply histogram equalization before normalization. Histogram equalization is a technique that allows us to improve the contrast of images and it has been found to be a powerful technique in image enhancement. Equalized histogram of a discrete gray-level image represents the frequency of occurrence of all gray-levels in the image and well distributes the pixels intensity over the full intensity range. 
 Finally, to prevent overfitting we used the same approach as \cite{Zagoruyko} and augmented training data applying affine transformation by rotating both patches in pairs to 90, 180, 270 degrees and flipping them horizontally and vertically.

\subsection{Network Architecture and Learning} \label{ssec:architectureAndLearning}

The proposed network architecture for one branch of the Siamese network of Fig.~\ref{fig:pipelineScheme} has following modules: convBlock[32,3,1,1]-convBlock[64,3,1,1]-pool[2]-convBlock[64,3,1,1]-convBlock[64,3,1,1]-pool[2]-convBlock[128,3,1,1]-
\noindent
convBlock\-[128,3,1,1]-pool[3]-convBlock[128,3,1,1]-L2norm. The shorthand notation: convBlock[\textit{N},\textit{w},\textit{s},\textit{p}] consists of a convolution layer with \textit{N} filters of size $\omega \times \omega$ with stride \textit{s} and padding \textit{p}, a regularisation layer (ReLU) and batch normalisation, pool[\textit{k}] is a  max-pooling layer of size $k \times k$ applied with stride \textit{k}. This architecture dubbed \textit{cnn7} was selected based on several experiments with different network structures having varying number of layers and involving also fully connected layers. We observed that convolutional networks without fully connected layers seemed to perform better than networks with fully connected layers, and \textit{cnn7} had the best performance among the networks we experimented.

\begin{figure}[t!]
  \begin{center}
   \includegraphics[scale=0.5]{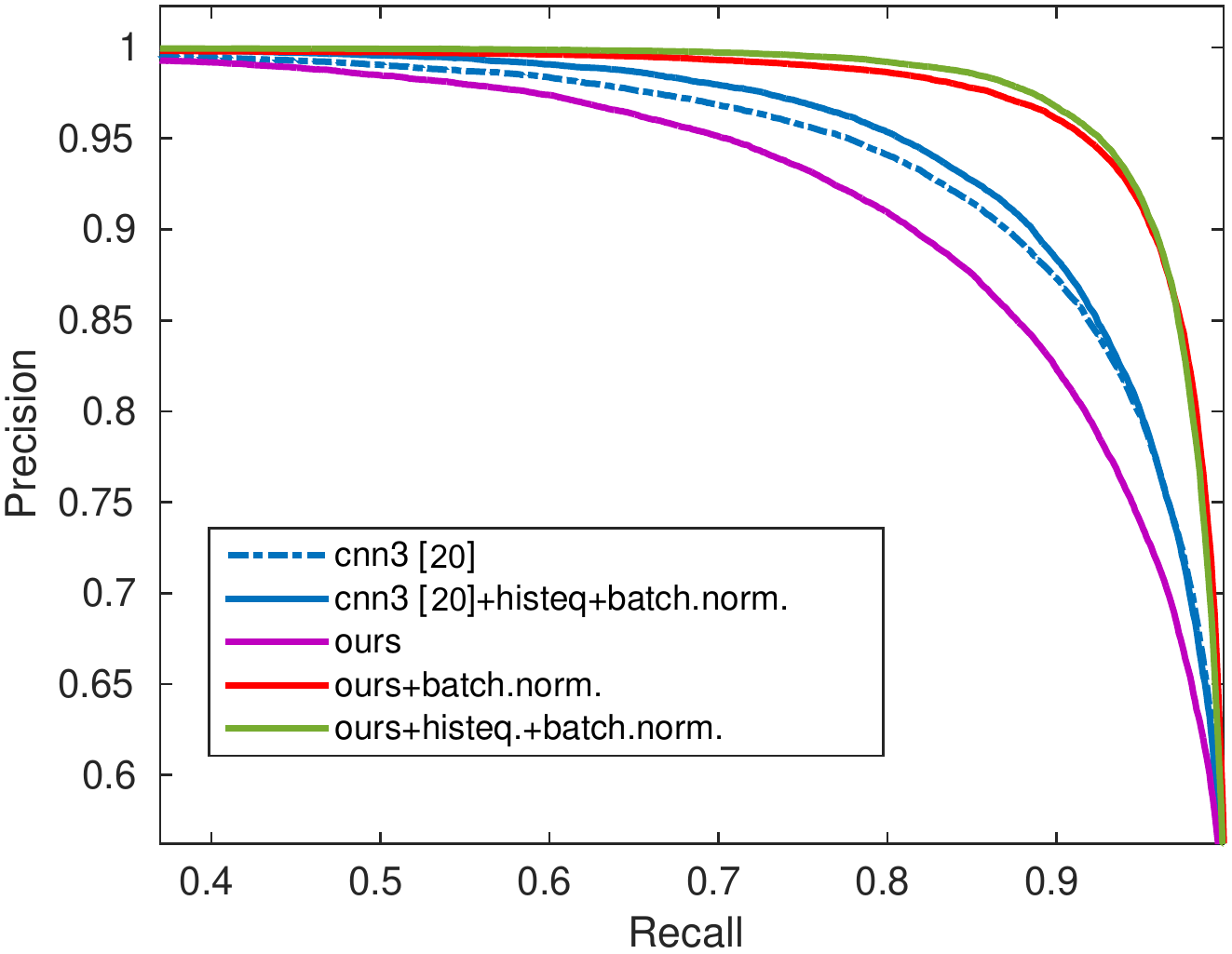}
  \end{center}
  \vspace{-3mm}
  \caption{Precision-recall curves for different descriptor architectures and data preprocessing approaches. To present the results more clearly we have zoomed on the recall axis. Here the performance is shown for 100k patch pairs of Notredame test data and the evaluated networks were trained in a Siamese architecture using 500k training pairs from the Liberty dataset.}
\label{fig:PRcurvesComparison}
\end{figure}

In our case, the benefit of applying batch normalization \cite{batchNorm} and histogram equalization was verified experimentally, as is shown in Fig.~\ref{fig:PRcurvesComparison} and described in Section 4. We also analyzed the network structure proposed by~\cite{simo2015deepdesc}, titled \textit{cnn3}, by re-implementing its architecture and utilizing contrastive loss objective function. 
As shown in Fig.~\ref{fig:PRcurvesComparison} we noticed that our network architecture clearly outperforms \textit{cnn3} even without histogram equalization of the input patches (blue and red curves respectively). Moreover, applying histogram equalization further improves the accuracy of the proposed method. 

In contrast to \textit{cnn3}~\cite{simo2015deepdesc} model and two models \textit{siam-$l_2$}, \textit{pseudo-siam-$l_2$} proposed by \cite{Zagoruyko}, we decomposed convolutional layers with a big kernel size into several filters with smaller kernels ($3\times3$), separated by ReLU activations. According to \cite{Simonyan14c}, it increases nonlinearities of the whole network and makes the decision function more discriminative. Moreover, our model has only half the number of parameters compared to \cite{Zagoruyko}.

\paragraph{Learning.} We minimize Contrastive loss function~(\ref{eq:lossFunction}) over a training set using Stochastic Gradient Descent (SGD) with a standard back-propagation \cite{Backprop} and ADADELTA~\cite{Adadelta}. We train our descriptor in two stages. In the first stage, the training data has 500,000 patch pairs and it took about 1 day to finish 100,000 iterations of training, which is equal to 40 epochs of the training set. Weights are initialised randomly and the model is trained from scratch. In the second stage, we augmented the number of training samples up to 4M pairs by using also rotated and mirrored versions of the original patches, and then resumed training for another 20 epochs starting from pre-trained descriptor from the first stage. Learning rate (0.01), weight decay (0.001) as well as the size of mini-batch (100) remain constant during the training. The model\footnote[1]{Source code and the model will be made available upon publication.}
was trained using publicly available deep learning framework Caffe on one NVIDIA TITAN Z GPU.

\section{Experiments}\label{sec:experiments}

In this section, we present experimental results evaluating the proposed descriptor on MSC dataset. In order to compare results with previous work, we use exactly the same standard datasets for training and testing as used by e.g.\ \cite{BrownPatches,Zagoruyko}. That is, for each of the three subsets of MSC dataset (Liberty, Notredame, Yosemite) we use a test set of 100,000 pairs of patches originally provided by \cite{BrownPatches}. For training we utilize 500,000 pairs of patches from each subset (also provided by \cite{BrownPatches}). If we augment the training data by including rotated and mirrored versions of the original training patches, as described in Section 3, we get 4 million pairs from the original 0.5 million. We train three models by using training patches from the three different subsets, and evaluate each of the three models with test pairs of the two remaining subsets. In total we get six cases which are presented in Table~\ref{tab:performanceTable}.

\begin{table*}[b!]
  \caption{Performance comparison of our descriptor and existing methods on Liberty (Lib), Notredame (ND) and Yosemite (Yos) image patches of MSC dataset. Numbers are false positive rate at 95\% recall on each of the six combinations of training and test sets. \textbf{Bold} numbers are the best across all algorithms. The proposed architecture can outperform \cite{Zagoruyko} in 4 cases out 6 and has the lowest average errors $mean$ and $mean_{[1:4]}$ (the average over the first four columns) with histogram equalization.} \label{tab:performanceTable}
\vspace{-3mm}
  \begin{center}
   \resizebox{1\textwidth}{!}{%
    \begin{tabular}{ c  c | c  c  c  c  c  c  c | c  c }
    \hline
     \multirow{2}{*}{Method}  & \multirow{2}{*}{Dim} & Training & \multicolumn{2}{c}{Yos}& \multicolumn{2}{c}{ND} & \multicolumn{2}{c|}{Lib} & \multirow{2}{*}{$mean$} & \multirow{2}{*}{$mean_{[1:4]}$}\\
      & & Test & Lib & ND & Lib & Yos & ND & Yos \\\hline
      nSIFT + L2 (no training) & 128d & & 29.84 & 22.53 & 29.84 & 27.29 & 22.53 & 27.29 & 26.55 & 27.38 \\
      \thead{nSIFT squared diff.\\ 
      linearSVM} 			  & 128d & & 27.07 & 19.87 & 26.54 & 24.71 & 19.65 & 25.12 & 23.83 & 24.55 \\
      Brown et al w/PCA        & 29d  & & 18.27 & 11.98 & 16.85 & \textbf{13.55} & - & - & -   & 15.16 \\
       \thead{Zagoruyko siam-L2~\cite{Zagoruyko},\\
      4M training pairs}                 & 256d & & 17.25 & 8.38 & 13.24 & 15.89 & \textbf{6.01} & 19.91 & 13.45 & 13.69 \\
      \thead{Zagoruyko pseudo-siam-L2~\cite{Zagoruyko},\\
      4M training pairs}			      & 256d & & 18.37  & 8.95 & 16.58 & 15.62 & 6.58 & 17.83 & 13.99 & 14.88 \\\hline
      Ours, 500k training pairs   & 128d & & \textbf{14.88} & 9.47 & 16.57 & 19.50 & 9.01 & 17.21 & 14.44 & 15.11\\
      Ours, 4M training pairs     & 128d & & 15.48 & 8.88 & \textbf{11.84} & 17.78 & 8.40 & \textbf{15.07} & 12.91 & 13.50\\
      \thead{Ours, 500k training pairs, \\
      hist. eq.}  & 128d & & 15.32 & 9.10 & 12.82 & 15.52 & 8.63 & 17.05 & 13.07 & 13.19\\
      \thead{Ours, 4M training pairs, \\
      hist. eq.}     & 128d & & 15.19 & \textbf{8.36}  & 12.20 & 14.72 & 6.93  & 15.86  & \textbf{12.21} & \textbf{12.74} \\\hline
     \end{tabular}
     }
  \end{center}
  \label{tbl:performanceComparison}
\end{table*}

\paragraph{Performance Metric.} We follow the standard protocol of  \cite{BrownPatches} and  calculate ROC curves by thresholding the distance between feature pairs and determine the false positive rate at 95\% recall. The numbers are shown in Table~\ref{tab:performanceTable}. 
As in \cite{Zagoruyko}, we also report the $mean$ across all six combinations of training and test data. Like the original work \cite{BrownPatches}, we also provide $mean_{[1:4]}$ metric which is the mean across the four cases obtained by training models only on Yosemite and Notredame. 

\begin{figure}[t!]
\centering
	\begin{subfigure}{0.49\linewidth}
	\centering
		\includegraphics[scale=0.3]{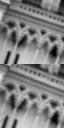}\hspace{0.1em}%
		\includegraphics[scale=0.3]{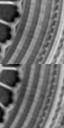}\hspace{0.1em}%
		\includegraphics[scale=0.3]{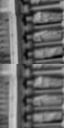}\hspace{0.1em}%
		\includegraphics[scale=0.3]{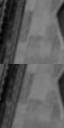}\hspace{0.1em}%
		\includegraphics[scale=0.3]{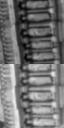}\hspace{0.1em}%
		\includegraphics[scale=0.3]{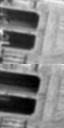}\hspace{0.1em}%
		\includegraphics[scale=0.3]{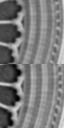}\hspace{0.1em}%
		\includegraphics[scale=0.3]{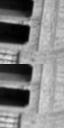}
		\caption{true positives}
	\end{subfigure}
	\begin{subfigure}{0.49\linewidth}
	\centering
		\includegraphics[scale=0.3]{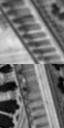}\hspace{0.1em}%
		\includegraphics[scale=0.3]{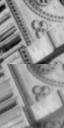}\hspace{0.1em}%
		\includegraphics[scale=0.3]{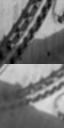}\hspace{0.1em}%
		\includegraphics[scale=0.3]{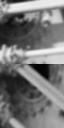}\hspace{0.1em}%
		\includegraphics[scale=0.3]{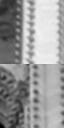}\hspace{0.1em}%
		\includegraphics[scale=0.3]{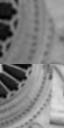}\hspace{0.1em}%
		\includegraphics[scale=0.3]{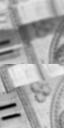}\hspace{0.1em}%
		\includegraphics[scale=0.3]{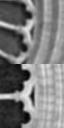}
		\caption{false negatives}
	\end{subfigure}\\
	\begin{subfigure}{0.49\linewidth}
	\centering
		\includegraphics[scale=0.3]{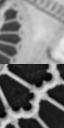}\hspace{0.1em}%
		\includegraphics[scale=0.3]{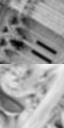}\hspace{0.1em}%
		\includegraphics[scale=0.3]{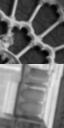}\hspace{0.1em}%
		\includegraphics[scale=0.3]{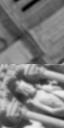}\hspace{0.1em}%
		\includegraphics[scale=0.3]{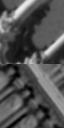}\hspace{0.1em}%
		\includegraphics[scale=0.3]{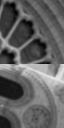}\hspace{0.1em}%
		\includegraphics[scale=0.3]{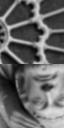}\hspace{0.1em}%
		\includegraphics[scale=0.3]{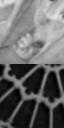}
		\caption{true negatives}
	\end{subfigure}
	\begin{subfigure}{0.49\linewidth}
	\centering
	\includegraphics[scale=0.3]{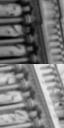}\hspace{0.1em}%
		\includegraphics[scale=0.3]{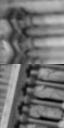}\hspace{0.1em}%
		\includegraphics[scale=0.3]{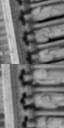}\hspace{0.1em}%
		\includegraphics[scale=0.3]{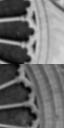}\hspace{0.1em}%
		\includegraphics[scale=0.3]{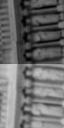}\hspace{0.1em}%
		\includegraphics[scale=0.3]{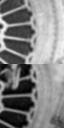}\hspace{0.1em}%
		\includegraphics[scale=0.3]{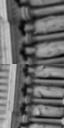}\hspace{0.1em}%
		\includegraphics[scale=0.3]{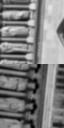}
		\caption{false positives}
	\end{subfigure}
\caption{Top-ranking true and false matches of Notredame patches by our best model.} \label{fig:tpfntnfp_examples}
\end{figure}

The results in Table~\ref{tab:performanceTable} confirm that the proposed model has better performance than~\cite{Zagoruyko} with the same number of training pairs. For instance, in Notredame-Liberty \textit{siam-L2} outperforms hand-crafted descriptor nSIFT+L2 and nSIFT squared diff. by 16.6\% and 13.3\% respectively in absolute error. Our method with the same size of training data further improves accuracy by 1.4\% in absolute error rate. Moreover, the length of our descriptor is significantly shorter than in \cite{Zagoruyko}. The benefit of applying histogram equalization is presented in the last two rows of Table~\ref{tab:performanceTable}.  The proposed model achieves 12.21\% and 13.07\% average error for with and without augmentation of training data, respectively. In general, it improves the performance of the proposed descriptor by 9.21\% in relative units for average error and by 6.93\% for $mean_{[1:4]}$ compared to \cite{Zagoruyko}.

\begin{table}[b!]
    \caption{Performance (area under precision-recall curve) of our descriptor architecture and \textit{cnn3} proposed by~\cite{simo2015deepdesc} on MSC dataset for 500k training pairs. Precision-recall curves corresponding to Liberty (Lib) training data and Notredame (ND) test data for considered descriptors are also illustrated in Fig.~\ref{fig:PRcurvesComparison}}
	\begin{center}
		\begin{tabular}{ c | c  c  c  c  c  c}
    			\hline
	    		 {Descriptor} & \multicolumn{2}{c}{\small Yos}& \multicolumn{2}{c}{\small ND} & \multicolumn{2}{c}{\small Lib}\\
		     {architecture} & {\small Lib} & {\small ND} & {\small Lib} & {\small Yos} & {\small ND} & {\small Yos} \\\hline
    			cnn3 & {0.943} & {\footnotesize 0.961} & {\footnotesize 0.950} & {\footnotesize 0.945} & {\footnotesize 0.964} & {\footnotesize 0.945} \\
	    		ours & {\footnotesize\textbf{0.977}} & {\footnotesize \textbf{0.984}} & {\footnotesize \textbf{0.980}} & {\footnotesize \textbf{0.977}} & {\footnotesize \textbf{0.985}} & {\footnotesize \textbf{0.975}} \\
	    \end{tabular}
	\end{center}
	 \label{tab:aucPerformance}
\end{table}

Fig.~\ref{fig:tpfntnfp_examples} shows top ranking false and correct matches of Notredame test dataset computed by our best model (the last row of Table~\ref{tbl:performanceComparison}). Specifically, we notice that some patches in false negative and false positive examples are so similar that even a human could make a mistake in interpretation. In fact, it seems that the top-ranking false positives (i.e.\ the pairs of negative patches whose descriptors are closest to each other) are probably originating from repeating texture patterns of the scene (i.e.\ similar texture appears in different 3D locations of the scene). Obviously, our descriptor or any other similar descriptor can not tell the difference here as it does not have access to multi-view information which was used to generate the ground truth labels.  More interestingly, the top-ranking false negatives (i.e.\ the pairs of positive patches with descriptors furthest away from each other) seem to originate from patches where there is a perceived dissimilarity because of inaccurate geometric alignment (due to non-planarity of the scene surface or due to inaccuracies in the orientation assignment or localization of the interest point). Thus, augmentation of training data and/or hard positive mining could bring further improvement and robustness to aforementioned factors in future. Nevertheless, Fig.~\ref{fig:tpfntnfp_examples} confirms the good behaviour of the proposed descriptor as the failure cases are intuitively understandable and hard to avoid in general without trade-offs.

Finally, we also calculated area under precision-recall curve for our method as this metric is used by \cite{simo2015deepdesc} for comparing descriptor performance. The results presented in Table~\ref{tab:aucPerformance} show that our network architecture performs better than the cnn3 architecture of \cite{simo2015deepdesc}.

\subsection{Spatial Transformer Networks}

Our visualisation in Fig.~\ref{fig:tpfntnfp_examples} shows that the image patches in many of the false negative pairs have a slightly differing alignment. That is, the patches represent corresponding scene surfaces but the scales, orientations and locations assigned by the interest point detector do not match precisely. Thus, based on the visualisation and interpretation of our results in Fig.~\ref{fig:tpfntnfp_examples}, we decided to further investigate that whether our descriptor could by made more robust to spatial misalignment by applying spatial transformer (ST) networks \cite{stn}. Specifically, the spatial transformer is a differentiable module performing explicit spatial transformations of input feature maps and can be placed at any part of a neural network easily. However, so far they have been mainly used in image classification problems \cite{stn} and, to the best of our knowledge, they have not been previously used for learning image similarity metrics with contrastive loss function.

\begin{figure}[h!]
\begin{center}
     \includegraphics[scale=.35]{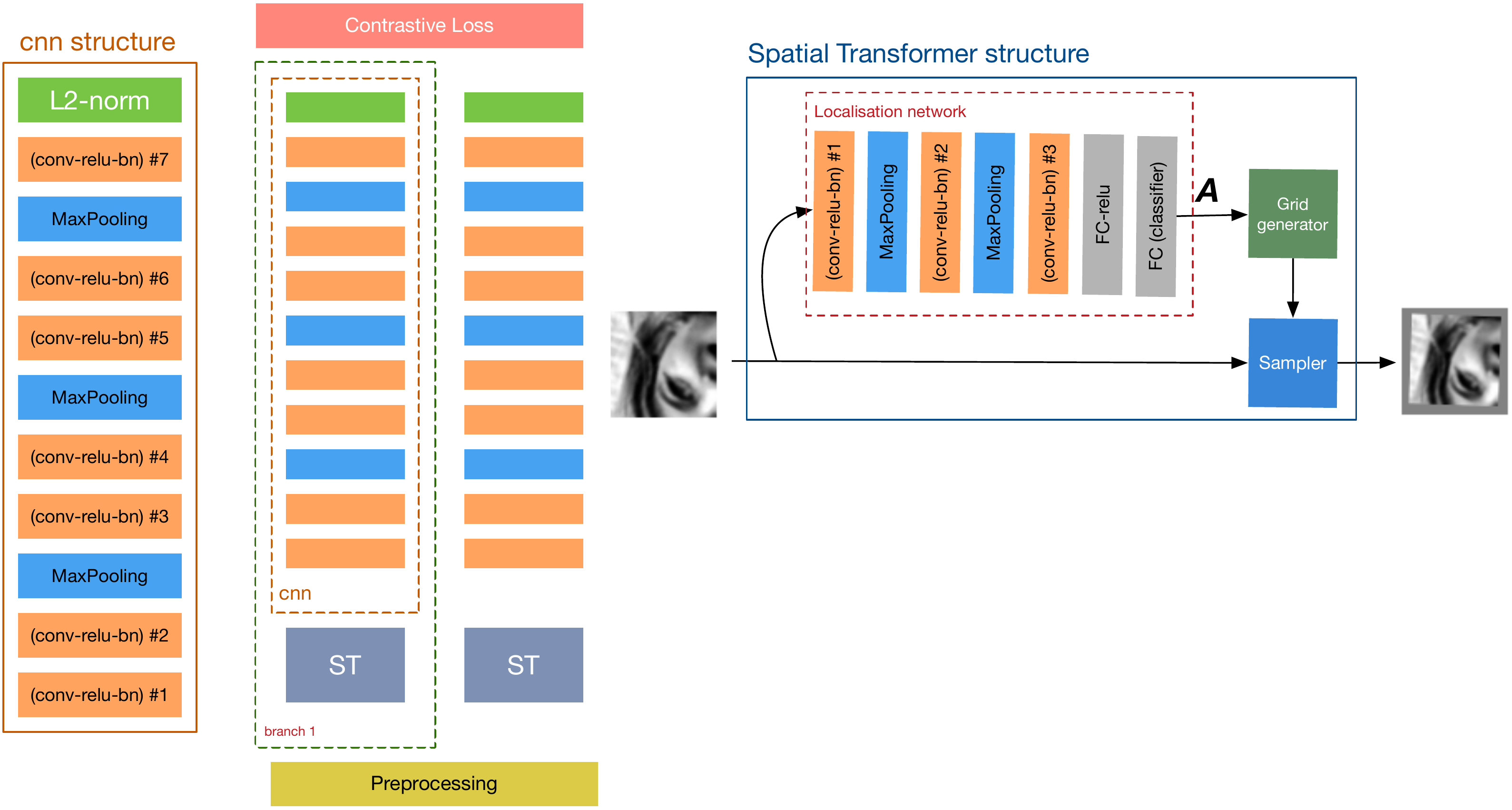}
     \caption{Schematic representation of the pipeline incorporating Spatial Transformer layer in our experiments. ST layer consists of three different parts: \textit{localisation network} predicts transformation parameters $\matr{A}$ that should be applied to the input feature map,  i.e.\ a raw input patch after preprocessing (histogram equalization) procedure. \textit{Grid generator} utilizes the predicted parameters $\matr{A}$ to construct a sampling grid which is used by \textit{sampler} to produce the transformed output. The size of both the input and output of ST layer is $64\times64$. The warped output of the spatial transformer is fed to CNN model respectively.}
\label{fig:stn_pipelineScheme}
\end{center}
\end{figure}

Fig.~\ref{fig:stn_pipelineScheme} schematically illustrates how we append our \textit{cnn7} model (introduced in Section 3.2) by incorporating ST modules right after the preprocessing layer.  As we put ST module as the first layer in the network, it directly transforms the preprocessed input patches. The number of parameters $\matr{A}$ can vary and depends on the type of transformation used. Inspired by examples of Fig.~\ref{fig:tpfntnfp_examples}, we aim to compensate errors caused by \textit{rotation}, \textit{translation} and \textit{scaling}. Therefore, the number of estimated parameters by localisation network equals 4 (one for rotation, one for scaling and two for translation transformations). The architecture of the localisation network is as follows: convBlock[32,5,1,2]-pool[2]-convBlock[64,5,1,2]-pool[2]-convBlock[128,5,1,2]-fc[256]-fc[4] where fc[\textit{n}] denotes a fully-connected layer with \textit{n} outputs. The complete model with the ST layer is denoted as \textit{cnn7stn}.

\begin{figure}[h!]
	\begin{subfigure}{0.48\linewidth}
		\centering
		\includegraphics[scale=0.43]{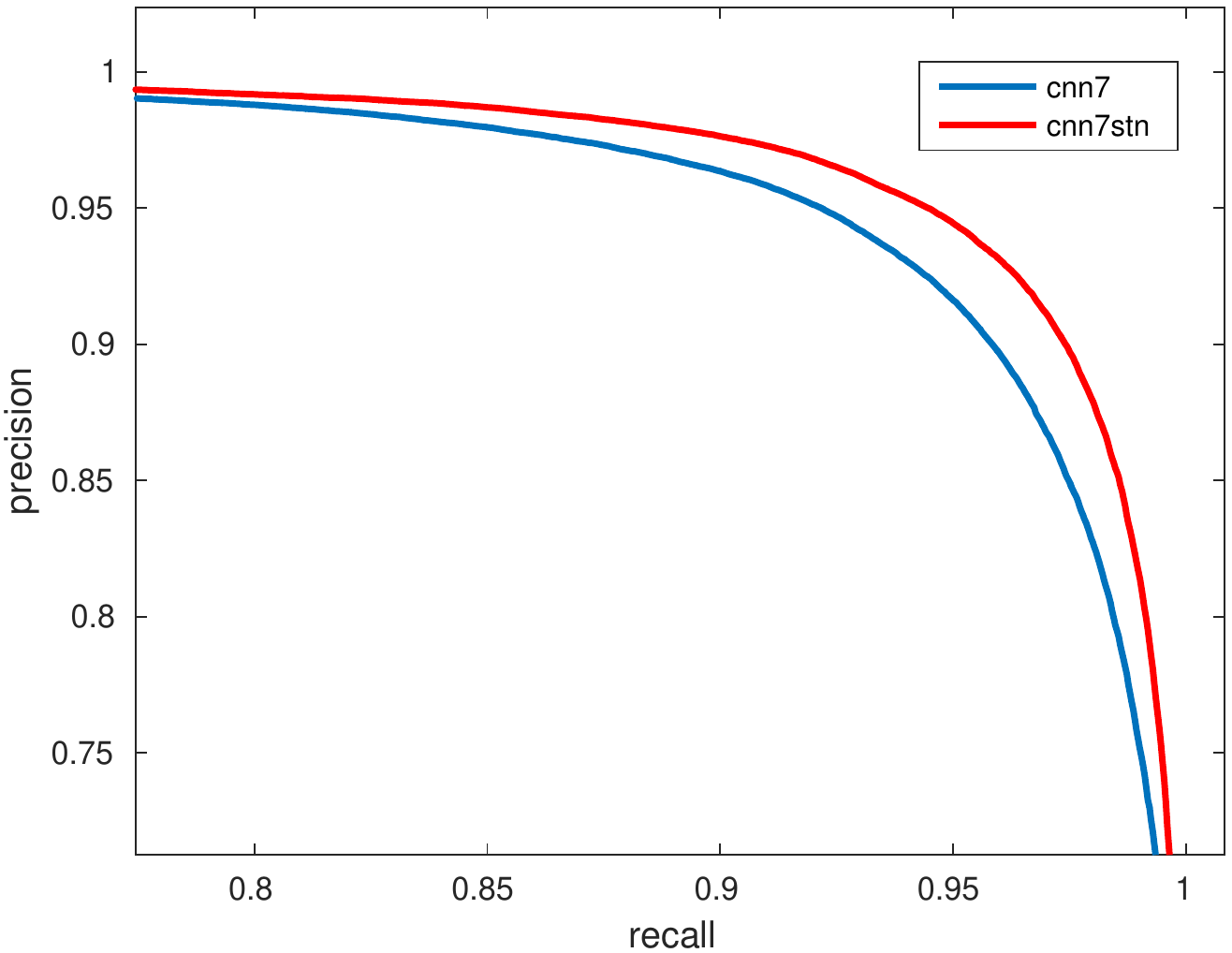}
		\caption{PR curve for training data}\label{sub:PR_train_stn_vs_base}
	\end{subfigure}
	\begin{subfigure}{0.48\linewidth} 
		\centering
		\includegraphics[scale=0.43]{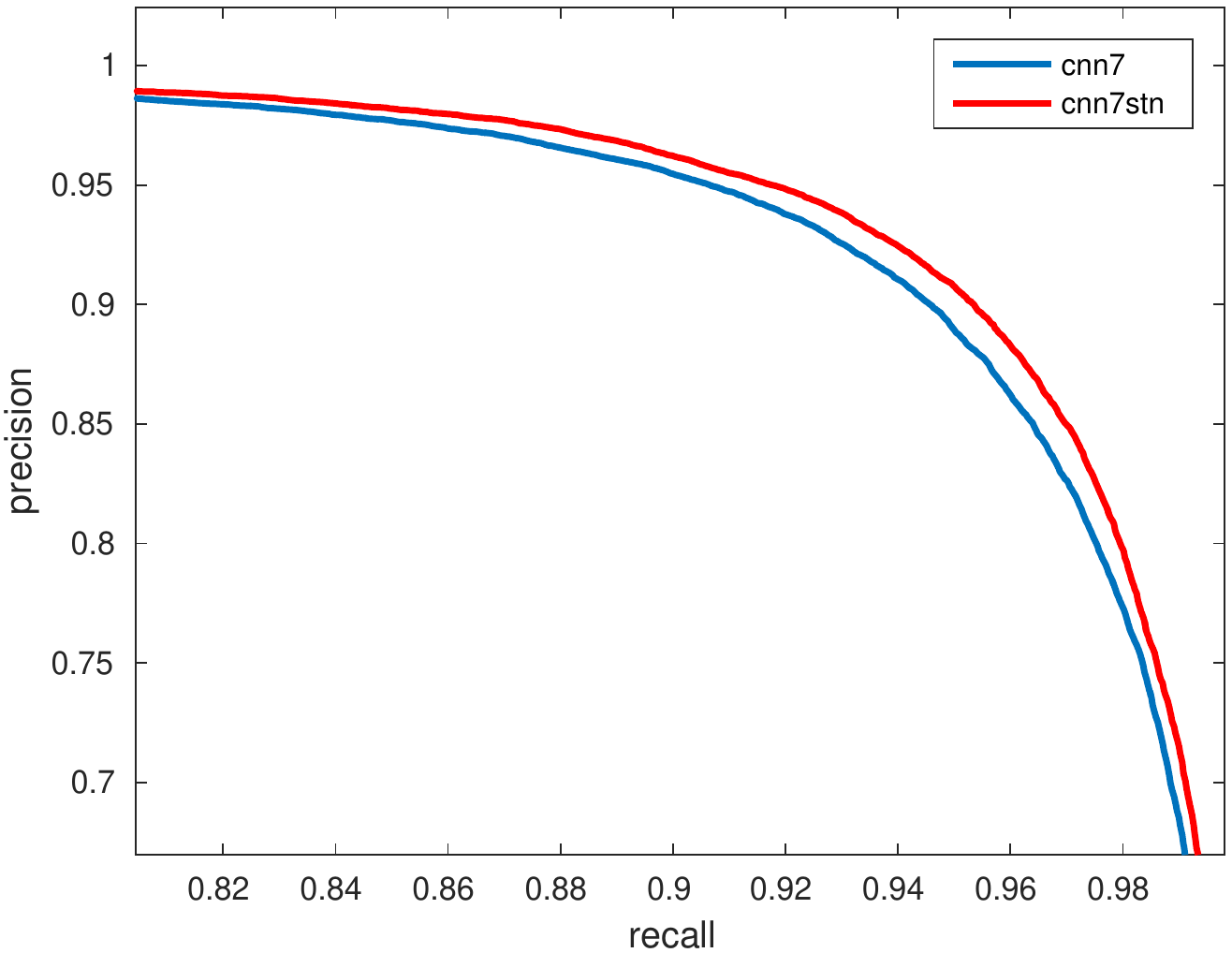}
		\caption{PR curve for test data}\label{sub:PR_test_stn_vs_base}
	\end{subfigure}
\caption{Precision-recall curves for training and test data for \textit{cnn7} and \textit{cnn7stn}.} \label{fig:stnResults}
\end{figure}

We train both \textit{cnn7} and \textit{cnn7stn} from random initialization using the histogram
equalized pairs from the augmented Liberty training set (4M pairs).
However, this time we did not use weight decay, and both models were trained
using a smaller number of epochs than used for the results of Table 1 (due to a
limited available training time). The models were evaluated with the NotreDame
test set (100k pairs) and the results are shown in Fig.~\ref{fig:stnResults}. We can see that \textit{cnn7stn}
gives better performance than \textit{cnn7}.

\begin{figure}[h!]
\centering
	\begin{subfigure}{\textwidth}
	\centering
		\includegraphics[scale=0.35]{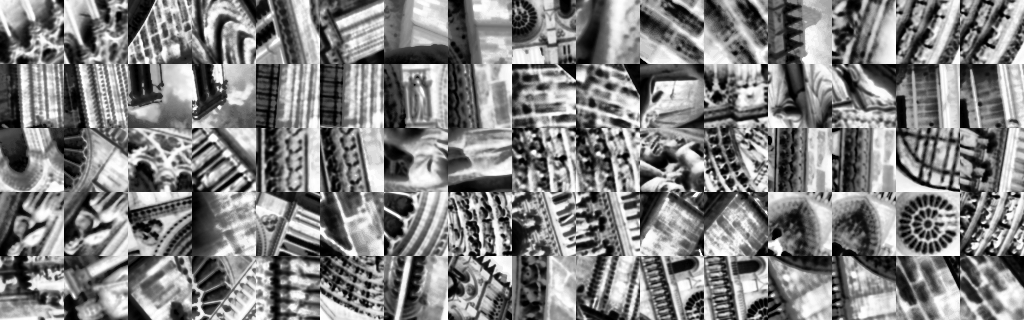}
		\caption{false negatives by cnn7 which are true positives by cnn7stn (in total 688 pairs)}
	\end{subfigure}
	\begin{subfigure}{\textwidth}
	\centering
		\includegraphics[scale=0.35]{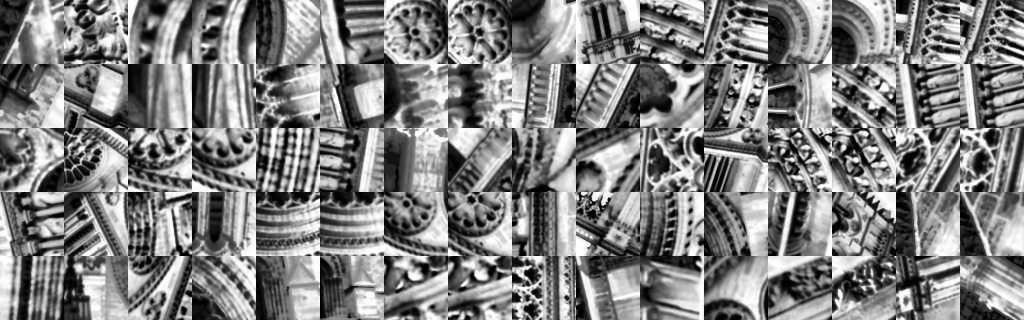}
		\caption{false positives by cnn7 which are true negatives by cnn7stn (2488 pairs)}
	\end{subfigure}
	\begin{subfigure}{\textwidth}
	\centering
		\includegraphics[scale=0.35]{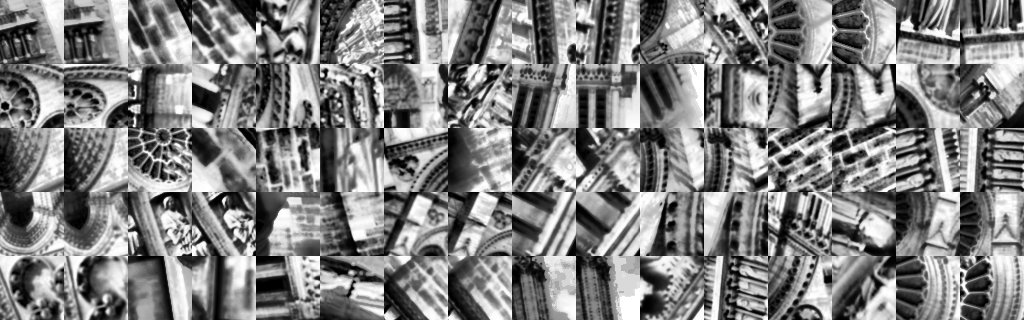}
		\caption{false negatives by cnn7stn which are true positives by cnn7 (688 pairs)}
	\end{subfigure}
	\begin{subfigure}{\textwidth}
	\centering
		\includegraphics[scale=0.35]{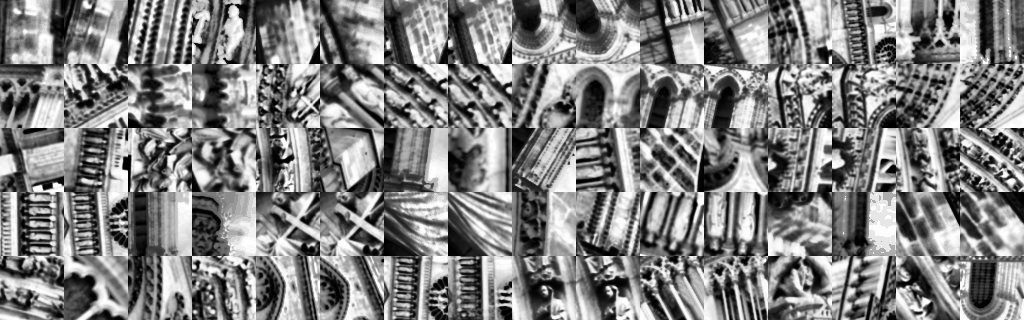}
		\caption{false positives by cnn7stn which are true negatives by cnn7 (1414 pairs)}
	\end{subfigure}
	\caption{Visualisation of some NotreDame test pairs which are classified differently by \textit{cnn7} and \textit{cnn7stn} when we set recall to 0.95 for both. As can be seen from Fig.~\ref{fig:stnResults} \textit{cnn7stn} has higher precision. The total number of test pairs is 100k. }\label{fig:stn_patches_performance_baseline} 
\end{figure}

\begin{figure}[h!]
\centering
	\begin{subfigure}{\textwidth}
	\centering
		\includegraphics[scale=0.35]{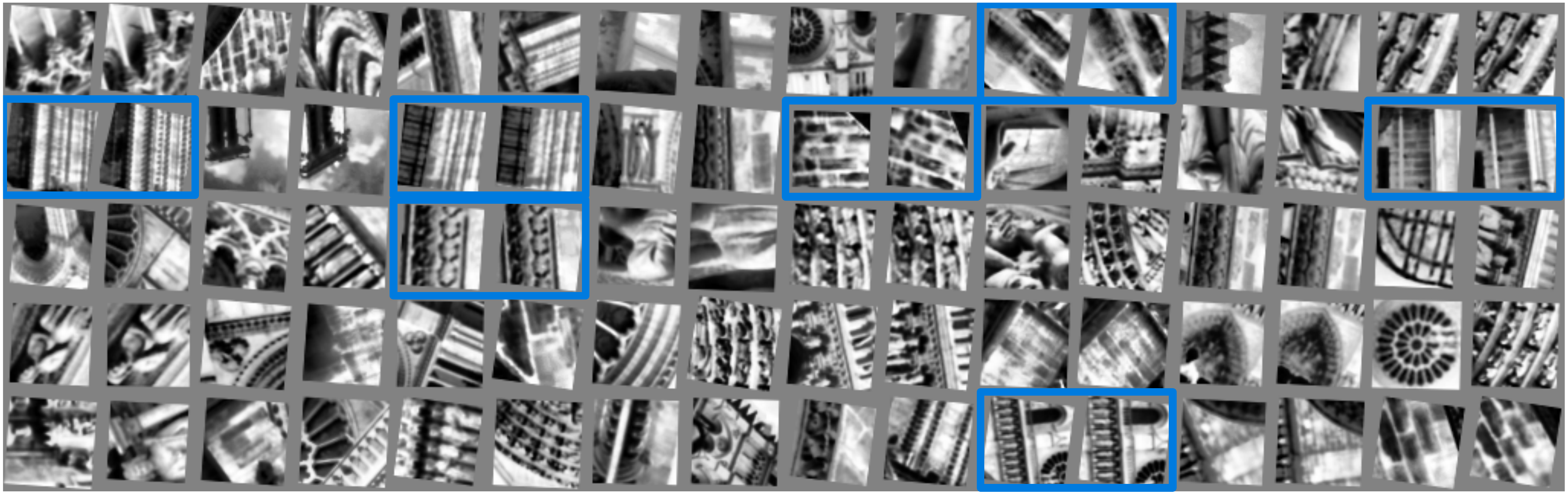}
		\caption{false negatives by cnn7 which are true positives by cnn7stn (688 pairs)}
	\end{subfigure}
	\begin{subfigure}{\textwidth}
	\centering
		\includegraphics[scale=0.35]{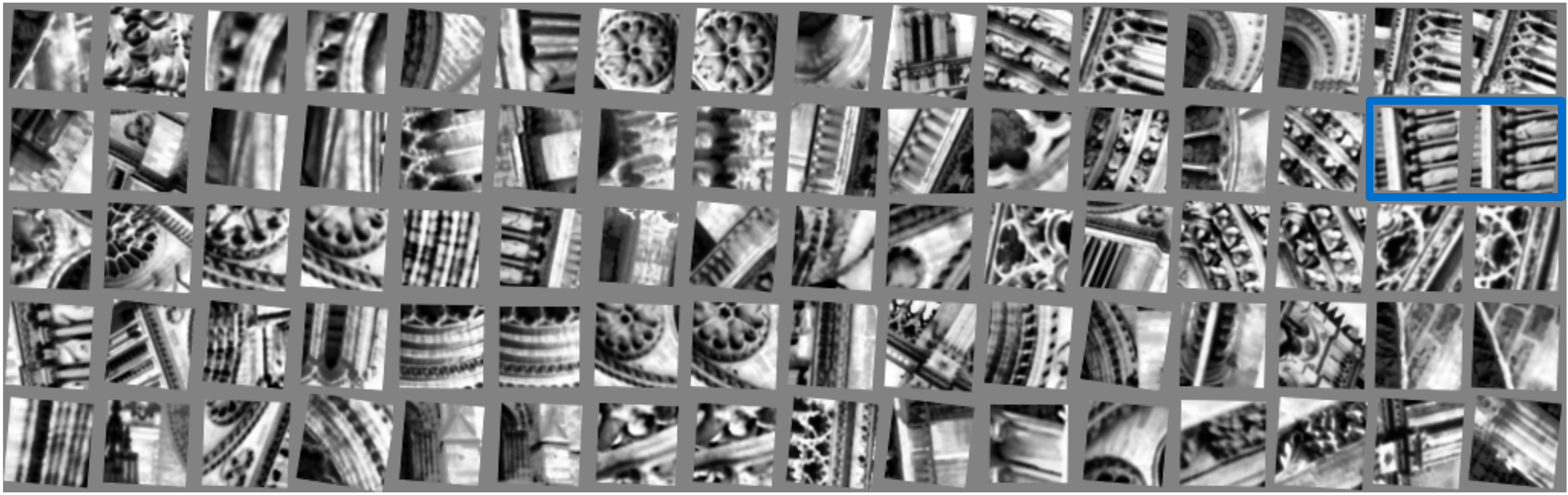}
		\caption{false positives by cnn7 which are true negatives by cnn7stn (2488 pairs)}
	\end{subfigure}
	\begin{subfigure}{\textwidth}
	\centering
		\includegraphics[scale=0.35]{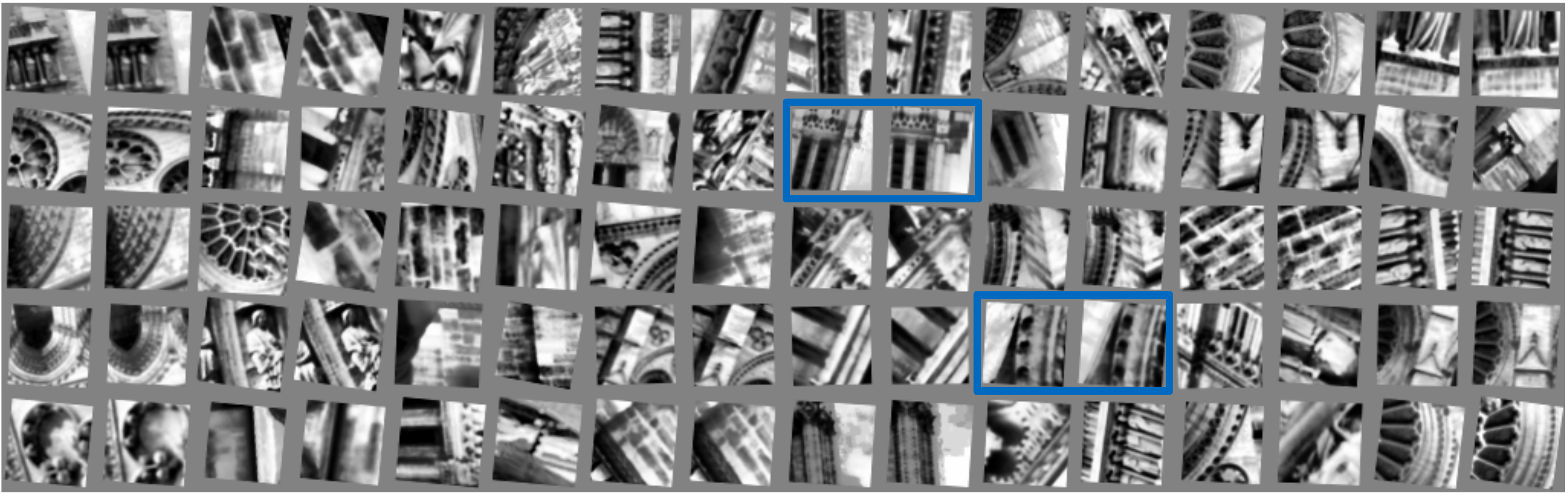}
		\caption{false negatives by cnn7stn which are true positives by cnn7 (688 pairs)}
	\end{subfigure}
	\begin{subfigure}{\textwidth}
	\centering
		\includegraphics[scale=0.35]{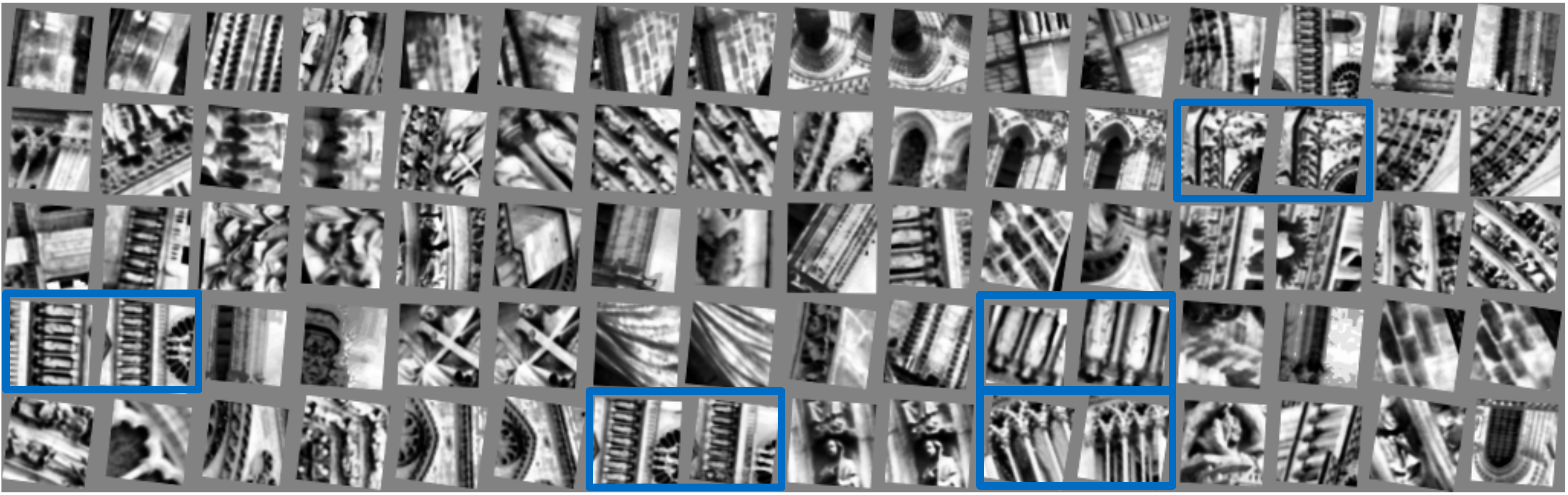}
		\caption{false positives by cnn7stn which are true negatives by cnn7 (1414 pairs)}
	\end{subfigure}
	\caption{The results of spatial transformations applied to the input image patches.
Blue color represents cases where ST layer transforms patches so that mutual
alignment is improved. }\label{fig:stn_patches_performance} 
\end{figure}

In order to further visualize and analyse the difference Fig.~\ref{fig:stn_patches_performance_baseline} shows examples of pairs for which the two models give different classification result at 0.95 recall. Fig.~\ref{fig:stn_patches_performance} shows the output of ST layer for the same patches. We can see that in most cases the ST layer transforms both patches of a pair quite similarly but in some cases (indicated with the blue color) the ST layer seems to improve the alignment which is probably the explanation for the better performance of \textit{cnn7stn}. Hence, it seems that the ST layer has learnt the desirable behaviour to some extent. Still, there is probably room for further improvements since many misaligned pairs remain quite differently aligned after the ST layer (cf.~Fig.~\ref{fig:stn_patches_performance}).

\section{Conclusion}\label{sec:conclusion}

In this paper, we use Siamese architecture to train a deep convolutional network for extracting descriptors from image patches. In training we utilized matching and non-matching pairs of image patches from MSC dataset. There are several conclusions that we can get from our experiments. First, we propose a descriptor with good performance, notably outperforming previous CNN-based $L_2$ norm descriptors on several datasets. We also show that utilizing histogram equalization for adjusting patch contrast improves the accuracy of the proposed model. In addition, we run preliminary experiments by appending our CNN architecture with spatial transformer layers and observe an improvement in the resulting descriptor. A potential future performance enhancement could be to investigate optimal structures of the localisation network of ST layers which could make the descriptor even more robust to geometric transformations.

\clearpage
\bibliographystyle{splncs}
\bibliography{accv}

\end{document}